\documentclass[conference]{IEEEtran}
\IEEEoverridecommandlockouts
\usepackage{cite}
\usepackage{amsmath,amssymb,amsfonts}
\usepackage{algorithmic}
\usepackage{graphicx}
\graphicspath{{figures/}}
\usepackage{textcomp}
\usepackage{xcolor}
\usepackage{lipsum}  
\ifCLASSOPTIONcompsoc
\usepackage[caption=false,font=normalsize,labelfon
t=sf,textfont=sf]{subfig}
\else
\usepackage[caption=false,font=footnotesize]{subfi
g}
\fi
\usepackage[colorlinks=true,linkcolor=black,anchorcolor=black,citecolor=black,filecolor=black,menucolor=black,runcolor=black,urlcolor=black]{hyperref}
\def\BibTeX{{\rm B\kern-.05em{\sc i\kern-.025em b}\kern-.08em
    T\kern-.1667em\lower.7ex\hbox{E}\kern-.125emX}}
\begin{document}

\title{A Trainable Sequence Learner that Learns and Recognizes Two-Input Sequence Patterns\\
\thanks{This report was produced in collaboration with the Institute of Neuroinformatics (University of Zurich and ETH Zurich)}
}

\author{\IEEEauthorblockN{Jan Hohenheim}
\IEEEauthorblockA{
\textit{University of Zurich}\\
Zurich, Switzerland \\
\href{mailto:jan@hohenheim.ch}{jan@hohenheim.ch} }
\and
\IEEEauthorblockN{Tommaso Stecconi}
\IEEEauthorblockA{
\textit{IBM Research GmbH}\\
\textit{Zurich Research Laboratory}\\
Work carried out at \textit{University of Zurich}\\
Zurich, Switzerland \\
tec@zurich.ibm.com}
\and
\IEEEauthorblockN{Zhaoyu Devon Liu}
\IEEEauthorblockA{\textit{CUHK BME}\\
\textit{Chinese University of Hong Kong}\\
Hong Kong, China \\
\href{mailto:zhyliu@link.cuhk.edu.hk}{zhyliu@link.cuhk.edu.hk} }
\and
\IEEEauthorblockN{Pietro Palopoli}
\IEEEauthorblockA{\textit{D-ITET}\\
\textit{ETH Zurich}\\
Zurich, Switzerland \\
ppalopoli@student.ethz.ch}
}

\maketitle

\begin{abstract}
We present two designs for an analog circuit that can learn to detect a temporal sequence of two inputs. The training phase is done by feeding the circuit with the desired sequence and, after the training is completed, each time the trained sequence is encountered again the circuit will emit a signal of correct recognition. Sequences are in the order of tens of nanoseconds. The first design can reset the trained sequence on runtime but assumes very strict timing of the inputs. The second design can only be trained once but is lenient in the input's timing.
\end{abstract}


\section{Introduction}
Sequence pattern recognition is both central to how our brain works and important for many modern AI applications such as speech recognition, speaker identification, automatic medical diagnosis or general classification. Since the human brain performs pattern learning and recognition with extreme energy efficiency, parallelism, and relatively good speed, the need to replicate these advantages in silica becomes apparent.

Some similar works already exist. Liu et al.\cite{Liu2021May} designed a multi-terminal transistor that can behave as a sequence detector by tuning the time delay between pulses fed into the transistor terminals. However, this system cannot train itself using certain input sequences like our brain.

As a starting point, we present an idea of how a simple sequence pattern learning and recognition chip could look like and provide two possible implementation schematics.

\section{Learning Algorithm}
Inputs are expected to be pulses coming from two different sources that arrive in some regular interval, determined by the design used. We will call them \emph{Signal A} (or just A) and \emph{Signal B} (or just B). They are instantly delayed by the same default amount of time. The goal is to make them overlap in time (learning phase) so that later the correct recognition can occur. This is done by treating A as a \emph{reference} and shifting B's delay up or down to make it coincide with A inside the circuit.

\begin{figure}[htbp]
\centerline{\includegraphics[width=200 pt,height=\textheight,keepaspectratio]{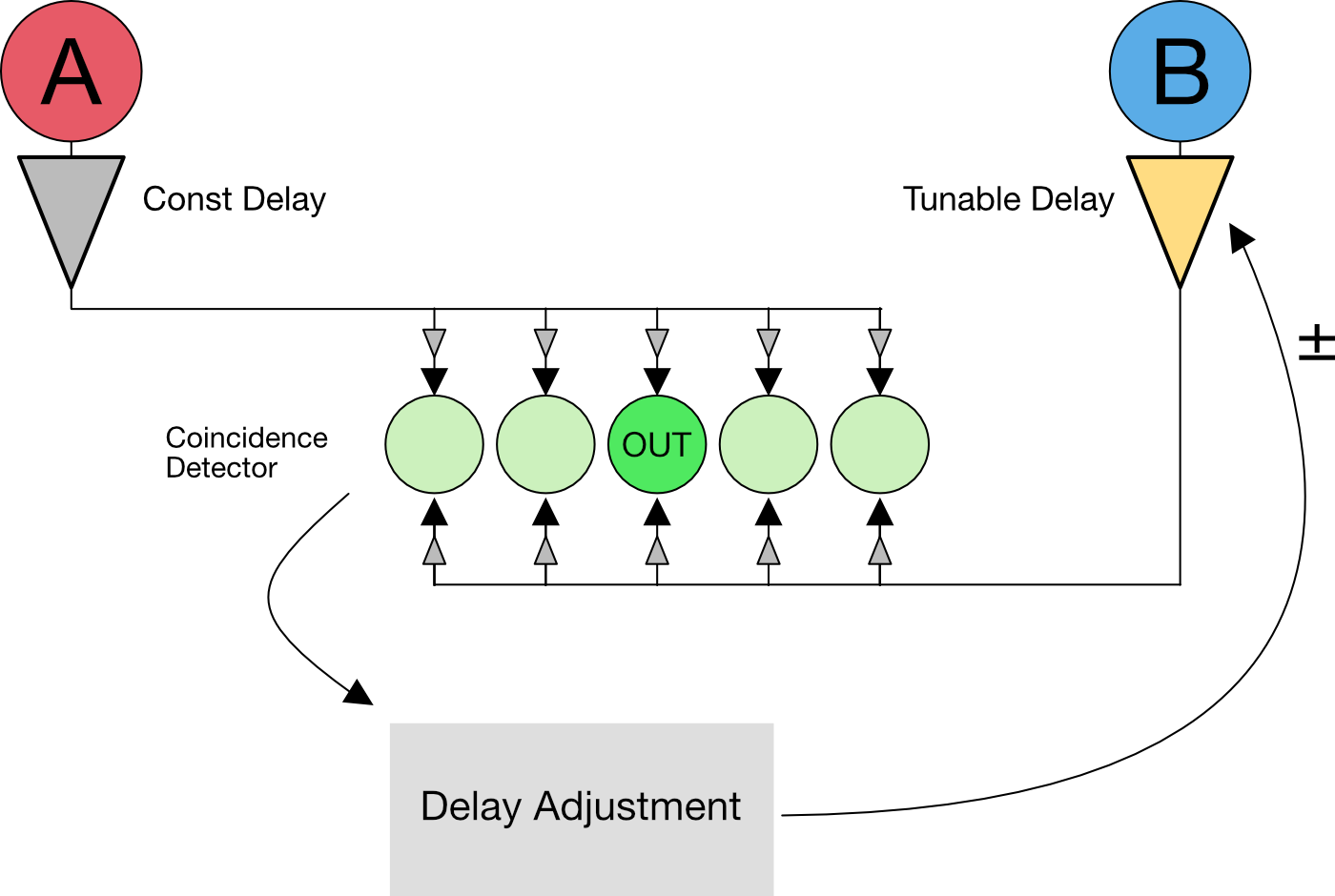}}
\caption{Overview of the shared logic behind the circuits}
\label{fig_logic}
\end{figure}

To do so, It is necessary to analyze the extent of the delay between the inputs, which is done by using a \emph{coincidence detector} inspired by the Jeffress model of sound localization \cite{Jeffress1948Feb}. This model works by letting the inputs move through a series of delaying nodes from opposite sides. They will meet at a node determined by their temporal offset. If they meet in the middle it means that they came in simultaneously. Since our inputs are assumed to always come with some offset, a meeting in the middle must mean that we successfully tweaked B's delay so that the training is finished. 

Before this is the case, the inputs will meet at another node, indicating how off from our goal we are. 
E.g. if we wish to detect the sequence ``AB'', B will come in after A, so the signals will meet on one of the nodes on the right in figure \ref{fig_logic}. 
meaning that we must aim to \emph{decrease} B's delay. This task is carried out by a \emph{delay adjustment unit} specific to the design in question. In any case, the delay modified is a row of 8 \emph{tunable delay}\cite{Lee2016Feb} units. Their delay is inversely proportional to a shared bias voltage.

\section{Sequence Learner Designs}
\subsection{Design A}

\begin{figure*}[!t]
\centering
\includegraphics[width=\textwidth, keepaspectratio]{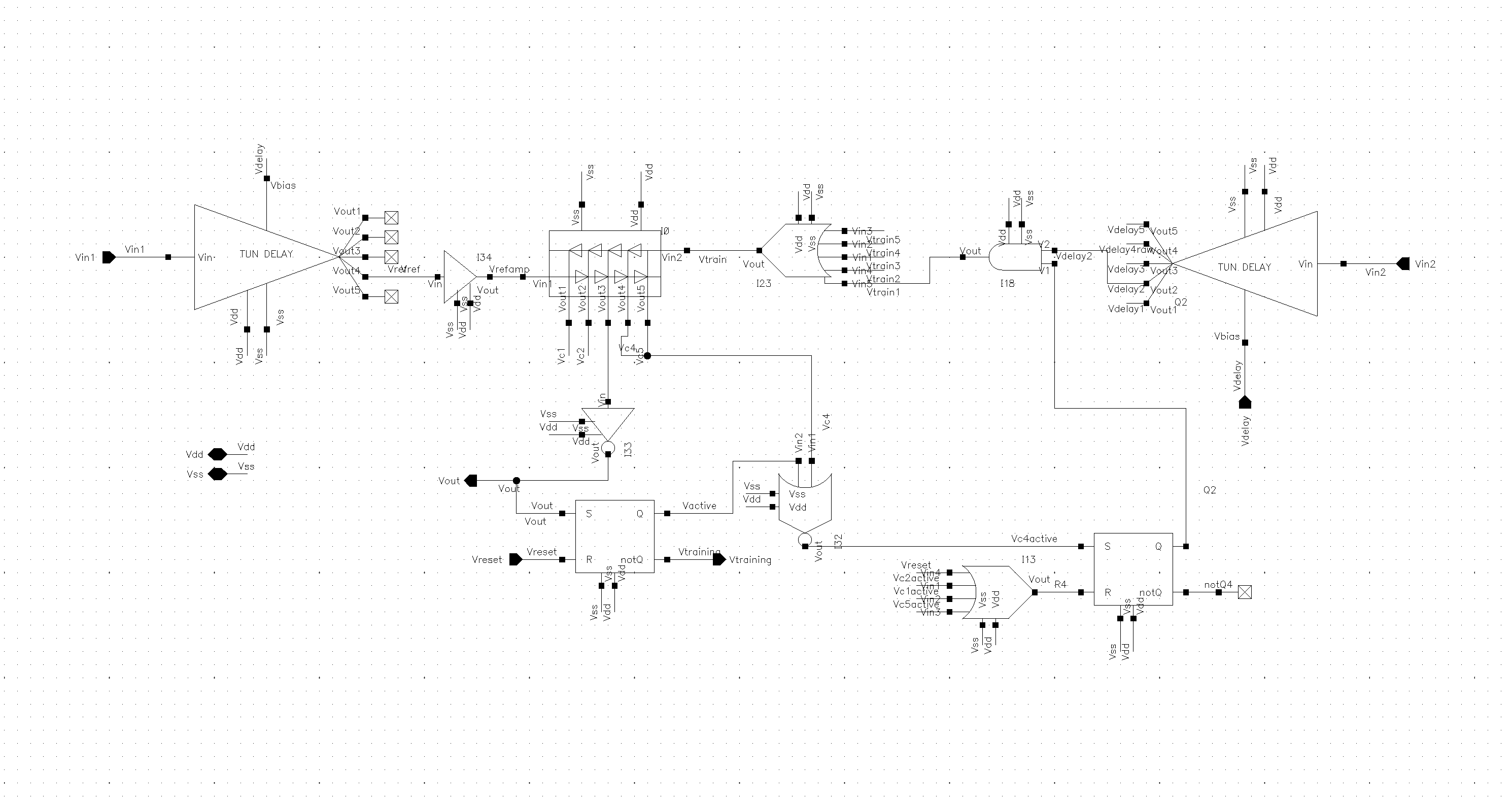}
\caption{A simplified view of the circuit's schematic. For clarity's sake, only the connections coming from the coincidence detector nodes in the middle and one to the right of the middle are included. In the full view, all nodes have equivalent connections}
\label{fig_schematic1}
\end{figure*}

\subsubsection{Usage}
\begin{description}[\IEEEsetlabelwidth{Auxiliary outputs}\IEEEusemathlabelsep] 
    \item[Main inputs] \(V_{\rm{in1}}\) and \(V_{\rm{in2}}\)
    \item[Auxiliary inputs] \(V_{\rm{reset}}\) to reset the learned sequence
    \item[Constant inputs] \(V_{\rm{delay}}\) at 1.18 V
    \item[Main outputs] \(V_{\rm{out}}\) for the detection
    \item[Auxiliary outputs] \(V_{\rm{training}}\) to tell a user that we are currently in training mode
    \item[Pulse length] 10 - 12 ns
    \item[Pulse delay] either 10 - 11 ns (so right after each other) or 20 - 21 ns
    \item[Pattern delay] at least 15 ns
\end{description}
Pattern delay is defined as the amount of time that needs to pass after a sequence finished until another one can be detected.

\subsubsection{Working principles}

The first version of the sequence learner uses the common design elements introduced above and SR latches as delay adjustment units (figure~\ref{fig_schematic1}). 

\begin{figure}[!t]
\centering
\includegraphics[width=2.5in, keepaspectratio]{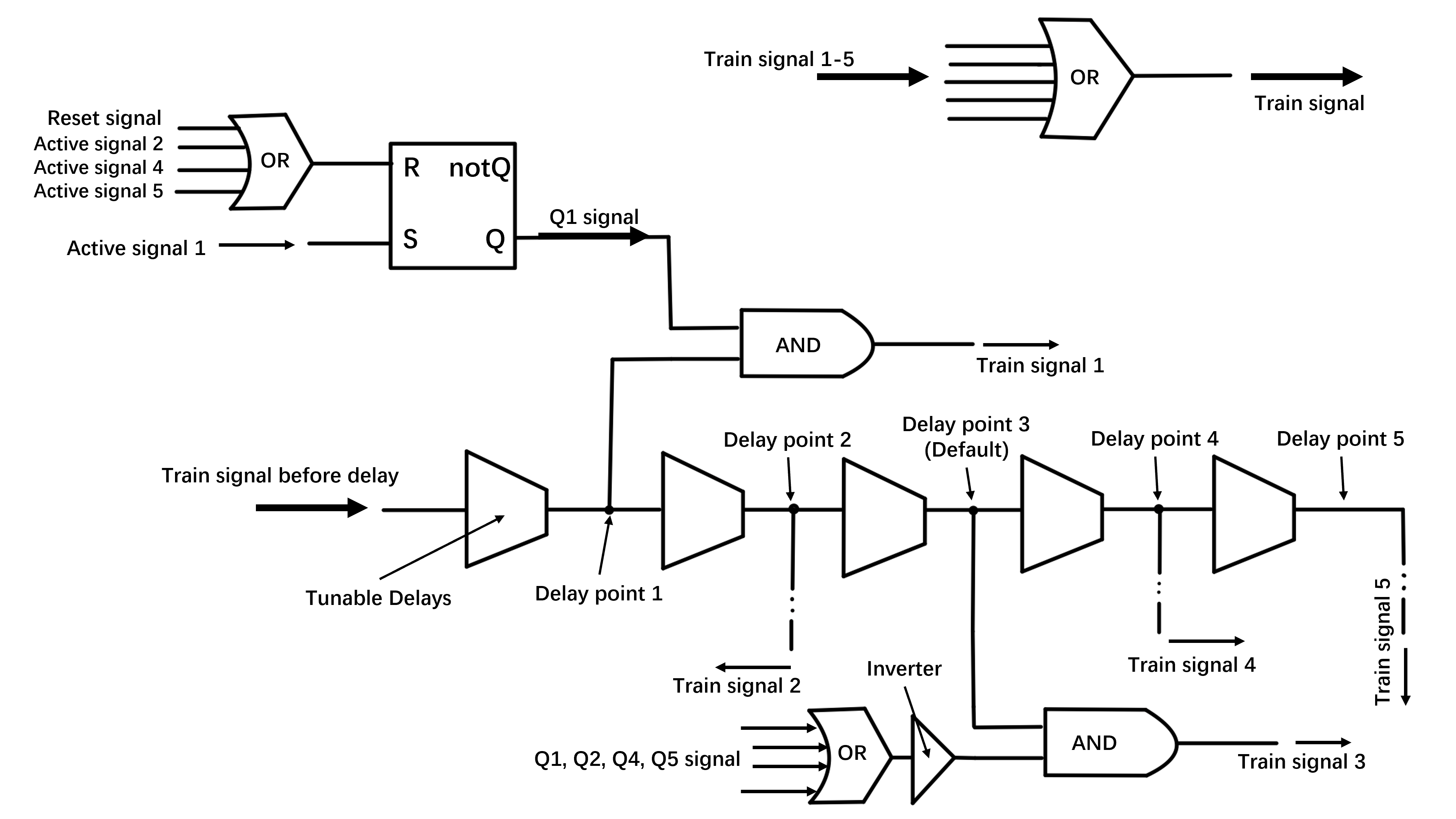}
\caption{A trained tunable delay unit. For simplicity, only the connections from the first subunit are included. If this SR latch is the only one active, the delay was successfully reduced.}
\label{fig_train}
\end{figure}

\begin{figure}[!t]
\centering
\includegraphics[width=2.5in, keepaspectratio]{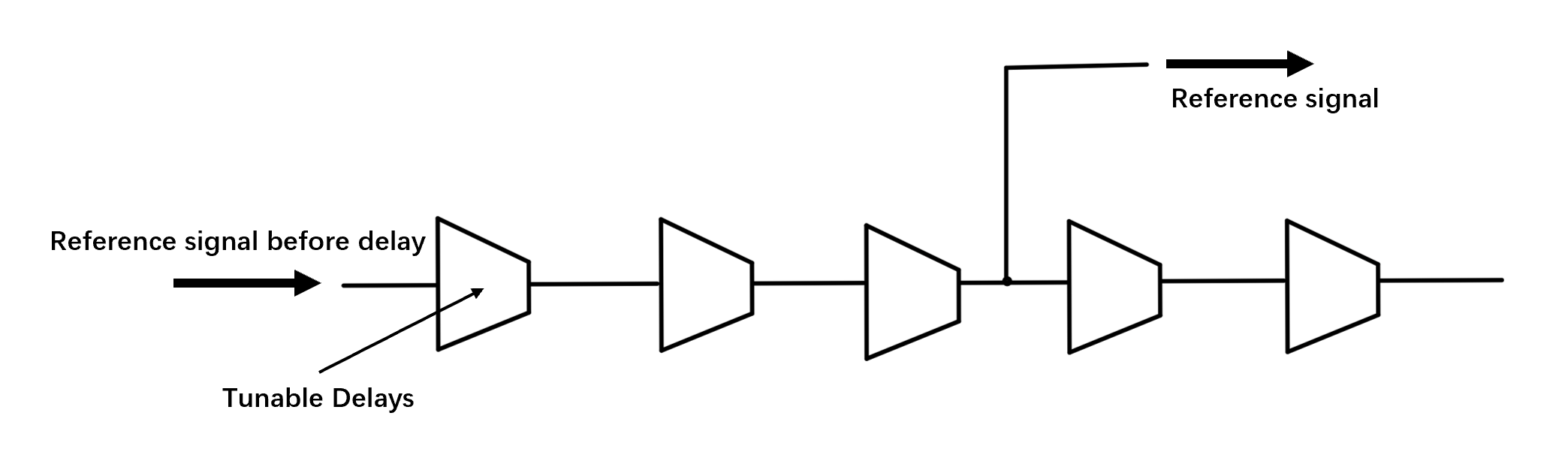}
\caption{The constant delay for the reference signal takes the input after it went through three tunable delay subunits.}
\label{fig_refer}
\end{figure}

Two pulses with a default delay are fed into two tunable delay lines. The signals are labeled as a reference signal and a trained signal, where the reference signal has a constant delay (figure~\ref{fig_refer}) while the trained signal will have its delay shifted (figure ~\ref{fig_train}).

In this design, the tunable delay line has a constant bias of 1.18 V, which makes every subelement have a delay of 5 ns, where one subelement consists of two tunable delay units. The delay shifting is not done by changing this bias (see the design B for that), but by reading the signal at different points along the delay line. This is done by ANDing the signals with SR latches (figure \ref{fig_train}). Only one of them is on at any time, ensuring that only one delay is selected. The outputs of all ANDs are ORd together to get the active signal.
The reference signal instead gets its constant delay by always picking up the signal after the third subunit (figure \ref{fig_refer}). 

The latches determining the delay for the train signal get set by the coincidence detector. Its output is a NAND between every pair of nodes, so an output of 0 means that a coincidence was found\footnotemark. If this did not happen in the middle, the training is either not finished or it is finished and we didn't find the right sequence. To differentiate between these, there is an SR latch controlling whether or not we are in \emph{training} or in \emph{active} mode via its output \(V_{\rm{active}}\). Its inverse is \(V_{\rm{training}}\) and it can be reset via \(V_{\rm{reset}}\). It is set when we detect the first coincidence in the middle of the detector.

\footnotetext{An earlier design kept the analog nature of the coincidence detector by using a bump-antibump \cite{Delbruck1991Jul} instead of NANDs. This led to multiple coincidences being detected at the same time, which we tried to remedy using a hysteretic winner-takes-all \cite{Indiveri2001Sep}. All of this ended up producing roughly the same results as just using digital NANDs, so we use them now instead.}

\begin{figure}[!t]
\centering
\includegraphics[width=2.5in, keepaspectratio]{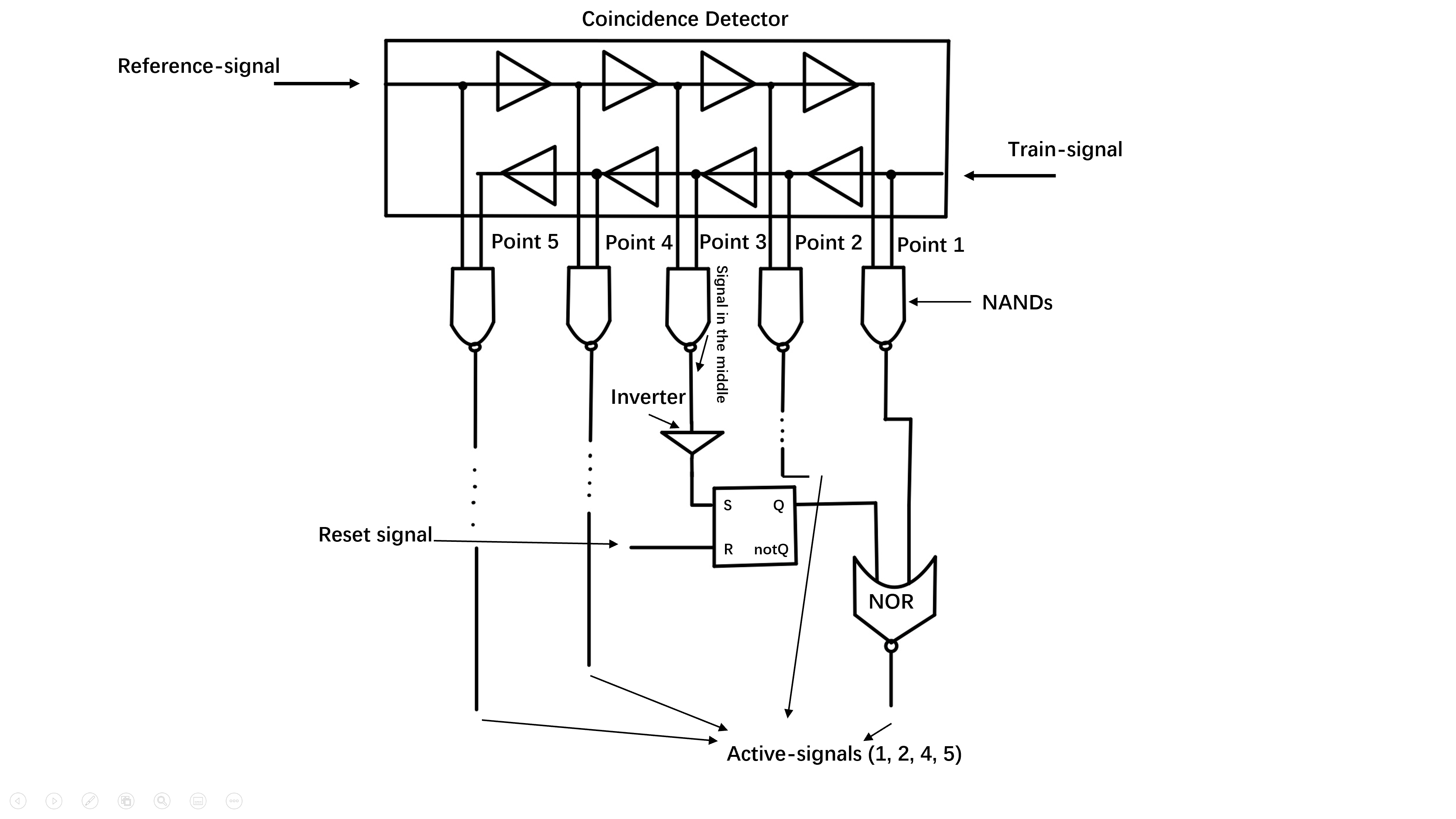}
\caption{The diagram of the coincidence detector fed with delayed Train-Signal and Reference-Signal. The ellipsis in the diagram indicates the same way of connection as between point 1 and point 3.}
\label{fig_cd}
\end{figure}

To ensure we are only updating the trained delay while actually training, we NOR the output of the coincidence detector with \(V_{\rm{active}}\) (figure \ref{fig_cd}). This output then is only \(V_{\rm{dd}}\) when both \(V_{\rm{active}}\) and the coincidence detector's output are 0. This output will both set the SR latch that allows reading the correct delay and reset all other latches.

\subsection{Design B}

\subsubsection{Usage}
\begin{description}[\IEEEsetlabelwidth{Auxiliary outputs}\IEEEusemathlabelsep] 
    \item[Main inputs] \(V_{\rm{in1}}\) , \(V_{\rm{in2}}\) 
    \item[Constant inputs] \(V_{\rm{default\ delay}}\) and \(V_{\rm{clock}}\) (values described later)
    \item[Main outputs] \(V_{\rm{out}}\) for the detection
    
    \item[Pulse length] 10 - 15  ns
    \item[Pulse delay] 10 - 50 ns 
    \item[Pattern delay] at least 15 ns
\end{description}

\subsubsection{Working principles}

\begin{figure}[!t]
\centering
\includegraphics[width=200pt]{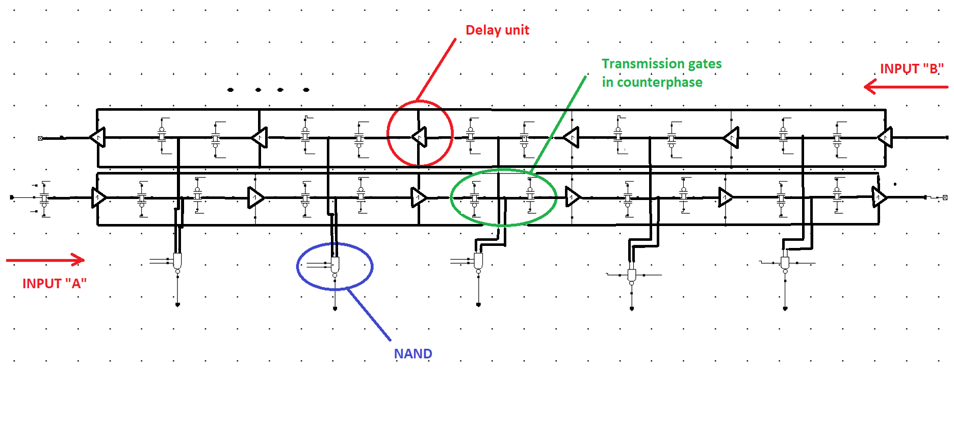}
\caption{The second version of the coincidence detector includes a sample and hold system made of two transmission gates in series working in antiphase.\\~\\}
\label{CD}
The transmission gates are controlled by a clock signal of period 10 ns, introducing a hold time \textgreater 5 ns. This solution improves the robustness of the coincidence detection and allows to work with ranges of delays rather than fixed ones.
\end{figure}

The second version of the sequence learner extends this functionality to flexible delays ranging between +- 50 ns (A with respect to B). Moreover, it internally provides the analog voltages required to tune the delays of the tunable delay element.
To extend the range of detectable delays we modified the coincidence detector block by including a pair of transmission gates working in antiphase between each stage of the coincidence detector (see figure \ref{CD}).

\begin{figure}[!t]
\centering
\includegraphics[width=200pt]{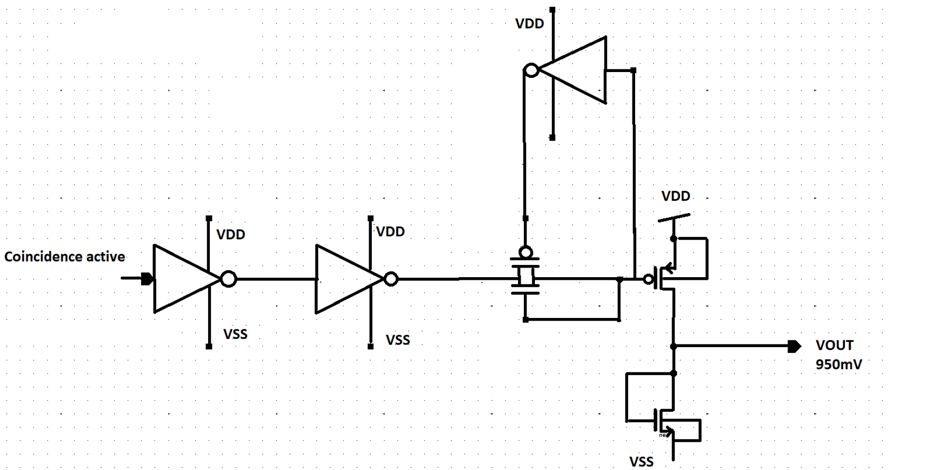}
\caption{The schematic for the circuit that converts the 20 ns delay output of the coincidence detector into the corresponding analog voltage driving the tunable delay element.\\~\\}
\label{950mV}
\end{figure}

The output of the coincidence detector can either activate a default delay of 60 ns, in case A and B are overlapped or 40 ns or 20 ns. In the latter two cases, a delay control logic is activated to disconnect the default input voltage of the tunable delay element of $\approx$ 700 mV.
The delay of 40 ns is given by an input voltage of $\approx$ 950 mV, which can be provided by an NMOS in transdiode configuration, mirroring its gate voltage, as shown in figure \ref{950mV}.

\begin{figure}[!t]
\centering
\includegraphics[width=200pt]{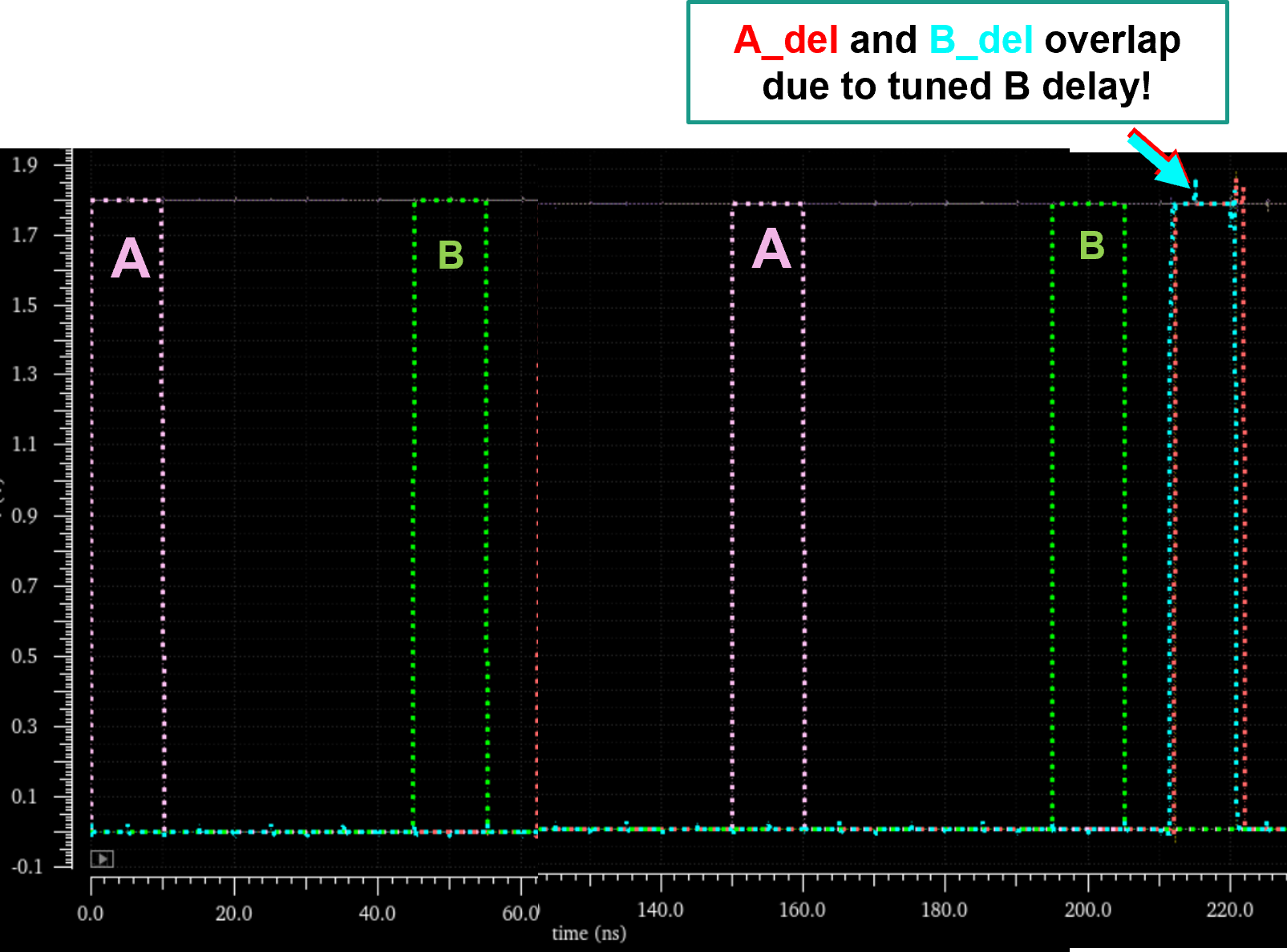}
\caption{Transient simulation of the sequence learner, when B is delayed by 45 ns.\\~\\}
\label{45ns}
\end{figure}

\begin{figure}[!t]
\centering
\includegraphics[width=175pt]{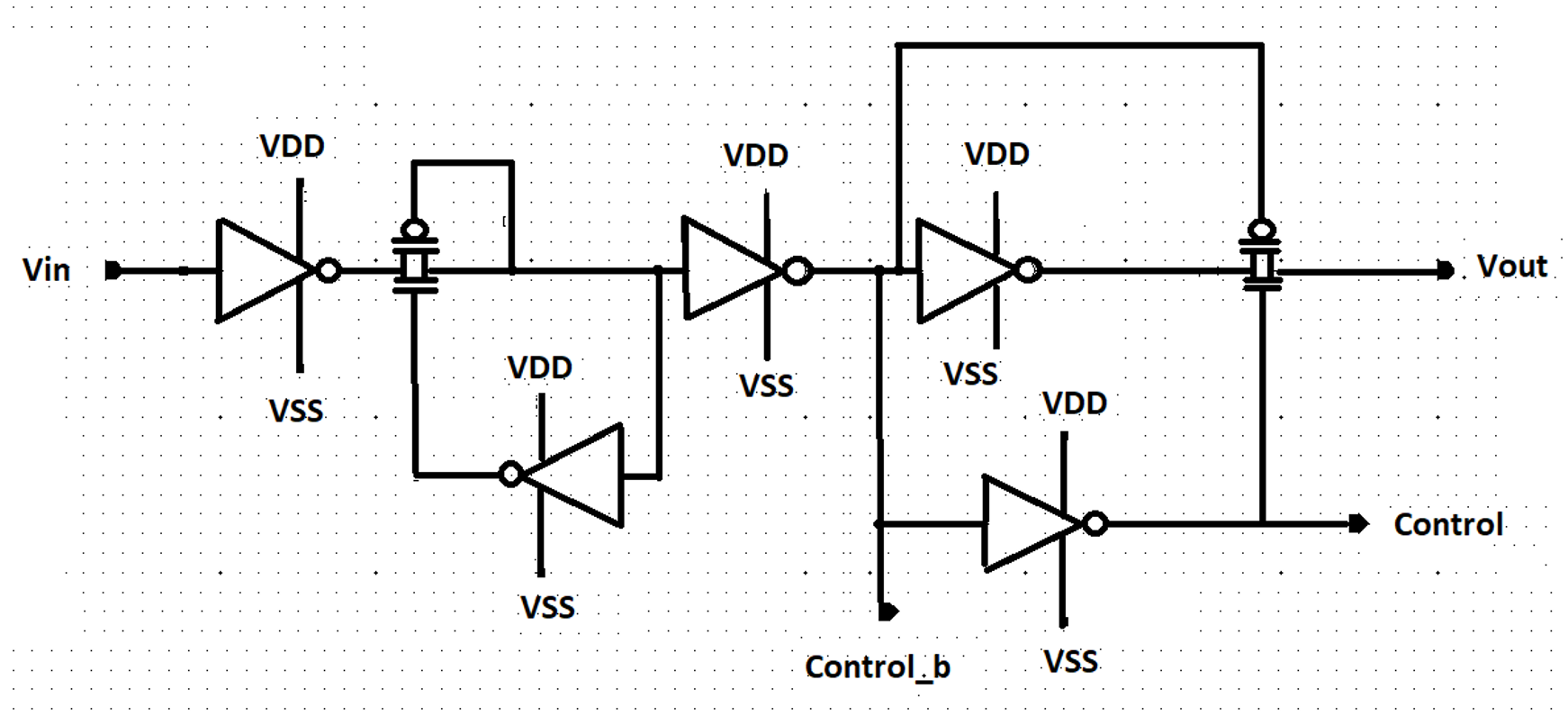}
\caption{Circuit schematics converting the 40 ns delay output of the coincidence detector into the corresponding 1.8 V driving the tunable delay element.\\~\\}
\label{CD_Force_18V}
\end{figure}

This solution can be used to set even longer delays; however, $\sim$ 700 mV correspond to a 60 ns delay and is already the threshold voltage for strong inversion regime. Therefore, for longer delays the input N-FET of the tunable delay must be biased in weak inversion, making the system highly sensitive to PVT and mismatches.\\~\\
We tested multiple sequence learning transient simulations, to evaluate the correct functionality of the circuit. The system can adjust the correct delay of 20 ns, for the range of delay 12 $\div$ 30 ns, and 40 ns for the range 32 $\div$ 50 ns, making the signals A and B overlap following the initial training. Only the specific minor ranges of 10 $\div$ 12 ns and 30 $\div$ 32 ns don't succeed, due to the signal distortion occurring in the tunable delay element (a 10 ns long input pulse gets distorted into an 8 ns long delayed output).\newline
In figure \ref{45ns} we show the transient simulation for a delay of B of 45 ns.
The other two main units implemented in this configuration  include logic for feeding the tunable delay with 1.8 V (as shown in figure \ref{CD_Force_18V}) and logic to deactivate the default voltage value in input to the tunable delay (which in our case is $\approx$ 700 mV) - shown in figure \ref{CD_Force_18V}.  
It is important to note that the latter circuit has two branches. The main branch connects directly to the tunable delay and provides 1.8 V . The second branch is used to deactivate the detection of 20 ns delay. This solution has been adopted because when a 40 ns delay detection occurs, an additional undesired 20 ns delay detection appears which must be suppressed.

\begin{figure}[!t]
\centering
\includegraphics[width=150pt]{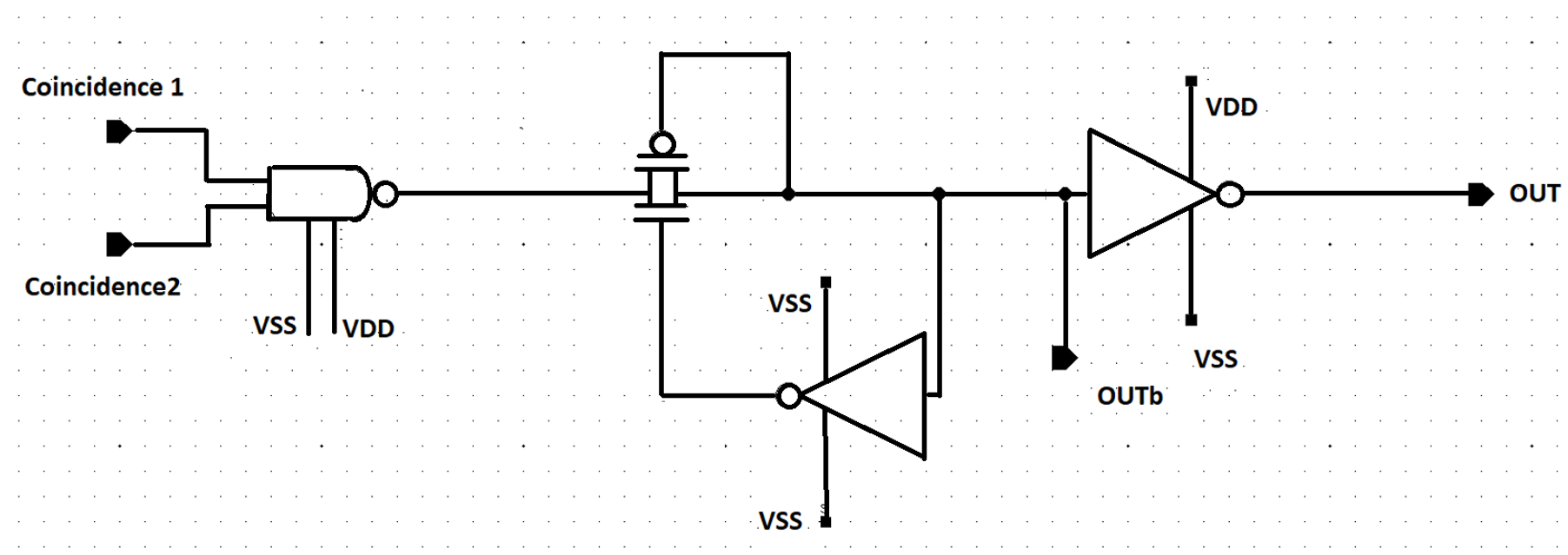}
\caption{Circuit schematics used to deactivate the pre-training default voltage in input to the tunable delay.\\~\\}
\label{CD_Deactivation}
\end{figure}

\begin{figure}[!t]
\centering
\includegraphics[width=175pt]{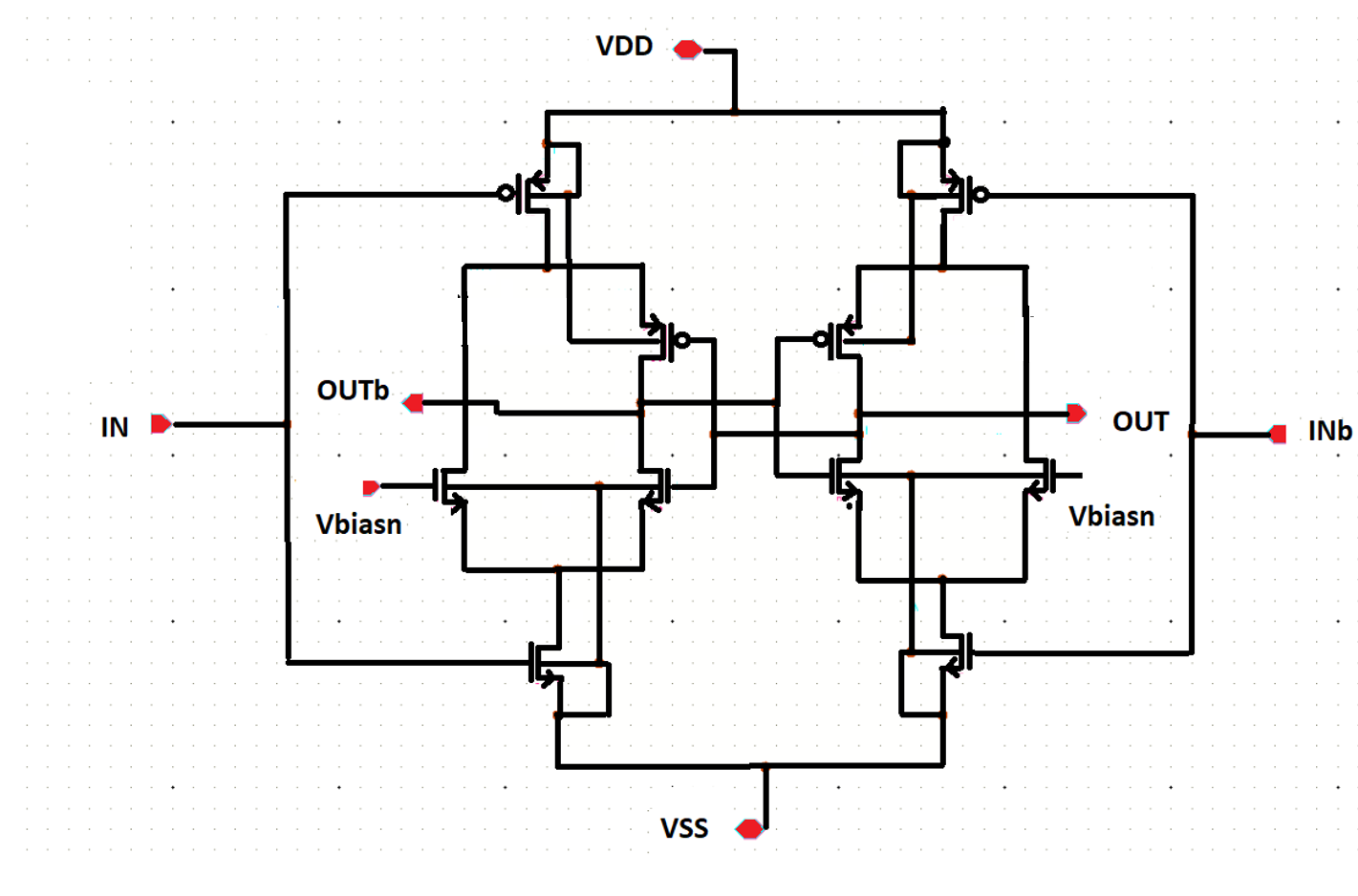}
\caption{Single unit of tunable delay obtained from [2]. 8 of these units in series build the tunable delay used in the sequence learner implementation. 
\\~\\}
\label{45ns2}
\end{figure}

\section{Figures of Merit}
\subsection{Design A}
    \begin{description}[\IEEEsetlabelwidth{Power consumption}\IEEEusemathlabelsep] 
    \item[Power consumption] static: negligible, dynamic: \(241\ \mu W\)
    \item[Operations] \(1.3 \times 10^7 \ s^{-1}\)
    \item[Area] \(2.1 \times 1.2\ mm\)
\end{description}
An operation is defined as either training a sequence, detecting a sequence or resetting the circuit.
The reported operations assume that it is impractical to let the inputs and outputs overlap. Since the output always comes with a fixed delay of 65 ns after the last input, a careful setup can feed as many inputs to the circuit as possible and let the inputs and outputs overlap, resulting in about 2.5 times more operations (\(3.1 \times 10^7\ s^{-1}\)).

The layout area was not prioritized at all and can massively be optimized without changing the circuit design.

\subsection{Design B}
\begin{description}[\IEEEsetlabelwidth{Power consumption}\IEEEusemathlabelsep] 
    \item[Power consumption] static: \(55\ \mu W\), dynamic: \(100\ \mu W\)
    \item[Operations] \(1.1 \times 10^7 \ s^{-1}\)
    \item[Area] Not layouted
\end{description}

\section{Applications and Modifications}
As is, the circuit can essentially just process a single bit: The incoming sequence is either AB or BA. The presented implementation can however be extended easily to support recognizing different delays between the signals.

This way, it could be used as a trainable directional detector for binaural auditory systems working in extremely high frequencies, e.g. learning that a mobile sound emitter is slightly to the left of the recording device and then detecting whenever the sound emitter returns to that location. The first design assumes pulse lengths and delays of just 10-20 ns. We can imagine this being of use for industrial robots that need to detect the position of other fast-moving machinery.

By altering the system a bit, one could in principle extend the operating range to support signals in the millisecond range, which would make it compatible with signals seen in biology. This comes closer to the original usage of the Jeffres model embedded within the circuits. This way, a circuit might be trained to detect each time a studied animal population visits a certain spot in conditions where a camera setup is not feasible, e.g. small mice in tall grass.

Another possible usage is to keep the circuit as a binary recognition machine that converts a signal coming from two sources into a single binary output. A sequence learner trained on ``AB'' would then convert the sequence ``ABABBAAB'' to ``1101''. In this interpretation, we reach a rate of 10 Mbit s\(^{-1}\). It is however not clear when such a thing would be useful.

The input range can also be extended to include any number of signals by adding nodes ``perpendicular'' to the coincidence detector. This could be used as a generic pattern detector. For example, every input could represent an individual letter and the sequence learner could recognize entire words.

Furthermore, detecting signals with a time-scale of nano-seconds is essential in some fields. 
R.Kamrla et. al\cite{Kamrla2022May} introduced the concept of 'delayed coincidence' and used 'delayed coincidence circuit' to distinguish between single photon spectra and pair electrons excited by a single photon. Anita et al.\cite{Anita2022Feb} introduced an anti-coincidence circuit with a refractory period to reduce the noise when detecting gamma-rays released by boron neutron. By implementing our circuit, it is possible to achieve the functions above.

\section*{Acknowledgment}
Our thanks to Prof. Shih-Chii Liu for her guidance and help during the long brainstorming sessions that went into the design of these circuits; Prof. Tobi for his advice and suggestions along the way; and the teaching assistants, classmates for their patient help and companionship in the course Neuromorphic Engineering by Institute of Neuroinformatics.

All four authors contributed equally to this paper.
\bibliography{Reference_conference}
\bibliographystyle{ieeetr}

\end{document}